\pdfoutput=1
\documentclass[preprint,12pt]{elsarticle}

\usepackage{amsmath,amssymb}
\usepackage{booktabs}
\usepackage{graphicx}
\usepackage{multirow}
\usepackage{subcaption}
\usepackage{siunitx}
\usepackage{float}
\usepackage{tabularx}
\usepackage{xcolor}
\usepackage{hyperref}
\hypersetup{
    hidelinks,
    pdftitle={A Unified Backbone--Expert Framework with Relation-Token and Residual--Classifier Interfaces for Automatic Modulation Recognition},
    pdfauthor={Zhixiang Deng, Houbiao Li, Zongyong Cui}
}

\begin{document}

\begin{frontmatter}

\title{A Unified Backbone--Expert Framework with Relation-Token and Residual--Classifier Interfaces for Automatic Modulation Recognition}

\author[uestc_math]{Zhixiang Deng}
\ead{m15181178952@163.com}
\author[uestc_math]{Houbiao Li\corref{cor1}}
\ead{lihoubiao0189@163.com}
\author[uestc_ice]{Zongyong Cui}
\ead{zycui@uestc.edu.cn}
\cortext[cor1]{Corresponding author.}
\affiliation[uestc_math]{organization={School of Mathematical Sciences, University of Electronic Science and Technology of China},
                    city={Chengdu},
                    postcode={611731},
                    country={China}}
\affiliation[uestc_ice]{organization={School of Information and Communication Engineering, University of Electronic Science and Technology of China},
                    city={Chengdu},
                    postcode={611731},
                    country={China}}

\begin{abstract}
Automatic modulation recognition (AMR) faces distinct representation bottlenecks under varying observation lengths, where a single model architecture often fails to excel. To address this, we propose a unified backbone-expert framework with a common convolutional state-space backbone and two specialized interfaces. For short sequences, we inject explicit lag-aware complex-plane descriptors as relation tokens before encoding to compensate for information loss. For long sequences, we design a gated multi-scale residual refinement module to correct the feature map, combined with a fixed-averaging classifier collaboration to harness complementary evidence. Our framework achieves overall average accuracies of $67.28\pm0.14\%$ on RML2016.10b and $87.19\pm0.77\%$ on HisarMod2019 (mean $\pm$ sample standard deviation over three runs), respectively. 
The framework's efficacy is further validated through three-seed ablations, native-length cross-configuration tests, and controlled window studies, confirming the benefit of expert-interface decoupling over one-size-fits-all architectures.
\end{abstract}


\end{frontmatter}

\section{Introduction}
\label{sec:introduction}

Automatic modulation recognition (AMR) identifies the modulation format of a received signal without transmitter-side metadata. It supports spectrum monitoring, cognitive radio, interference analysis, and non-cooperative communications. Early deep AMR systems learned representations directly from raw I/Q samples with convolutional networks~\cite{oshea2016convolutional}. Later work modeled temporal dependencies with recurrent and hybrid networks~\cite{hong2017rnn,west2017deep}. Other studies explored multimodal Transformers~\cite{shao2025iqformer}, lightweight complex-valued networks~\cite{xin2025cppcnet}, and mixture-of-experts classifiers~\cite{gao2026moe}.

Several recent models have been evaluated on both short- and long-observation benchmarks~\cite{shao2025iqformer,xin2025cppcnet}. These studies demonstrate broad applicability, but the available observation budget contributes to different representation bottlenecks. With limited temporal support, informative local relations can be difficult to recover under noise. Extended support provides richer evidence, but it also produces a longer feature sequence in which channel-distorted responses must be refined and complementary cues must be aggregated into one decision. Together with the signal and channel distribution of a benchmark, these differences motivate examining whether one generic expert form addresses both regimes effectively.

We therefore pair a common convolutional state-space backbone with two complementary expert designs. On RML2016.10b, lag-aware distance and complex-correlation descriptors supply explicit local complex-plane relations, and an early token interface allows the sequence encoder to contextualize them jointly with learned backbone features. On HisarMod2019, multi-scale residual refinement corrects the feature map over long receptive fields, while decision-level collaboration combines convolutional and state-space classification evidence. The resulting framework unifies the backbone and feature dimensions at the architectural level while assigning a corresponding expert inductive bias and integration interface to each evaluated benchmark condition.

The main contributions are as follows:
\begin{itemize}
    \item We propose a unified backbone--expert framework in which a common convolutional state-space topology supports two expert inductive biases and their corresponding integration interfaces.
    \item We design two complementary configurations. Relation-aware token augmentation supplies explicit local complex-plane structure before contextual encoding in the short regime. Multi-scale residual refinement and CNN--state-space decision collaboration support the long regime.
    \item We evaluate the framework with three-seed ablations, SNR-stratified analysis, native-length cross-configuration experiments, and controlled HisarMod2019 windows.
\end{itemize}

The remainder of this paper is organized as follows. Section~\ref{sec:related_work} reviews deep AMR and expert-integration methods. Section~\ref{sec:method} presents the proposed framework. Section~\ref{sec:experiments} describes the evaluation protocol, and Section~\ref{sec:results} reports the results. Section~\ref{sec:discussion} discusses the scope and limitations, followed by the conclusion in Section~\ref{sec:conclusion}.

\section{Related work}
\label{sec:related_work}

\subsection{Deep automatic modulation recognition}
Deep AMR replaces handcrafted decision statistics with representations learned from signal samples. Convolutional models established that local patterns can be learned directly from raw I/Q sequences~\cite{oshea2016convolutional}, while recurrent and convolutional--recurrent architectures introduced explicit temporal modeling~\cite{hong2017rnn,west2017deep}. A broader review by Zhang et al.~\cite{zhang2022survey} organizes these developments around neural architectures, input representations, benchmark datasets, complexity, and deployment challenges. These families remain strong baselines, but their local receptive fields or sequential recurrence can limit the efficient modeling of widely separated signal events.

Attention-based models address this limitation by modeling nonlocal dependencies. IQFormer combines convolution and staged Transformer blocks after dynamically embedding I/Q and time--frequency modalities~\cite{shao2025iqformer}. MST instead constructs parallel resolutions and exchanges information through cross-scale token fusion~\cite{zhang2025mst}. Graph-based methods offer another route: STF-GCN represents spatial, temporal, and frequency information as graph nodes and uses adaptive correlation to construct their connectivity~\cite{shao2025stfgcn}. These methods demonstrate the effectiveness of contextual and multi-domain modeling in their respective evaluation settings. However, they do not directly examine whether the same specialist representation and integration interface remain equally suitable across substantially different observation budgets.

Recent studies have also introduced selective state-space models into AMR. DWMTN combines parallel Mamba and Transformer branches using fusion weights predicted for each input~\cite{ma2025dwmtn}. ConvMamba combines convolution, Mamba2, and soft-threshold denoising for long-sequence high-order modulation recognition~\cite{zhu2025convmamba}. LM-GDMAF uses lightweight Mamba modules to extract temporal features and fuses I/Q and spectrogram modalities in a small-sample setting~\cite{zhang2026lmgdmaf}. These studies modify or combine sequence backbones to improve contextual modeling and feature fusion. The present work does not modify the selective state-space operator itself. Instead, it uses a common state-space sequence backbone to study how signal-specific expert information and its integration interface should be organized under the two evaluated benchmark conditions.

\subsection{Complex-valued and relation-aware signal representations}
The I/Q channels jointly describe a complex baseband signal. Processing them as unrelated real channels can therefore discard useful coupling. CPPCNet preserves this structure with complex partial pointwise convolution while controlling computational cost~\cite{xin2025cppcnet}. DualFormer forms separate I and Q token streams, encodes them with shared Transformer parameters, and fuses them at the prediction head~\cite{zhao2025dualformer}. Both methods respect complex signal structure, although they learn the relevant relations implicitly.

Explicit relations provide a complementary inductive bias. STF-GCN derives graph connectivity from correlations among multidomain features~\cite{shao2025stfgcn}. Multimodal systems instead combine I/Q samples with time--frequency maps or constellation diagrams~\cite{shao2025iqformer,guo2026adaptive}. These methods differ in what specialist information is constructed and where it enters the network. Our short path uses local lagged distance and complex correlation, then injects the resulting representation at the token interface. Unlike general multimodal fusion, it requires neither an additional signal transform nor an image modality.

\subsection{Feature refinement and complementary classification for long observations}
Long-observation AMR has been approached through signal restoration, multi-scale feature extraction, and hybrid sequence modeling. DAE-CNN-BiLSTM first applies a denoising autoencoder. Separate CNN and BiLSTM branches then extract and concatenate spatial and temporal features~\cite{long2025dae}. MAWDN instead learns an adaptive wavelet decomposition and aggregates decisions through a residual classification flow~\cite{qin2025mawdn}. Both approaches exploit complementary information, but the former reconstructs a cleaner waveform and the latter explicitly decomposes the signal into time--frequency scales.

Other models enlarge the effective context directly in the discriminative network. WACN combines local-window processing with depthwise and dilated convolution to balance local attention and receptive-field growth~\cite{feng2025wacn}, while DFENet uses parallel kernels to extract features at several temporal scales~\cite{le2026dfenet}. MILAFormer further combines convolutional feature enhancement, bidirectional recurrence, hierarchical attention, and Mamba-inspired linear attention~\cite{zhao2026milaformer}. These studies establish the value of denoising, multi-scale context, and complementary sequence representations. Our long configuration follows a different organization: it predicts a residual correction directly in the discriminative feature space and combines independently formed CNN and state-space decisions, without waveform reconstruction or an explicit transform-domain decomposition.

\subsection{Expert representations and integration interfaces}
Existing AMR fusion mechanisms operate at several interfaces. IQFormer performs dynamic fusion during token embedding~\cite{shao2025iqformer}. The adaptive multimodal framework of Guo et al. uses gated attention to combine features from three signal domains~\cite{guo2026adaptive} and DualFormer merges independently encoded I/Q streams near the prediction head~\cite{zhao2025dualformer}. MoE-AMC takes a different approach by assigning specialized experts to low- and high-SNR regimes through a learned gate~\cite{gao2026moe}. These studies establish that specialist representation, expert inductive bias, and integration location can all affect AMR performance.

Multi-scale sequence models are also related. MST fuses tokens produced by several convolutional resolutions within the same observation~\cite{zhang2025mst}, while ms-Mamba applies state-space blocks at different sampling rates to capture temporal structure efficiently~\cite{karadag2026msmamba}. Such designs enrich how a given observation is analyzed. Our framework instead starts from the representation bottleneck observed in each benchmark condition and assigns a corresponding expert and interface within one backbone template. Local relation tokens enrich evidence before contextual encoding in the short setting. Residual feature correction and classifier collaboration exploit the broader support available in the long setting. The framework therefore distinguishes relation-token augmentation from residual--classifier collaboration, rather than applying generic multi-scale processing throughout.

\section{Proposed method}
\label{sec:method}

\subsection{Problem formulation}
Consider a transmitted complex baseband sequence $s_m[\ell]$ whose modulation class is $m\in\mathcal{M}=\{1,\ldots,C\}$. A generic discrete-time received signal can be written as~\cite{zhang2022survey}
\begin{equation}
r[\ell]
=e^{j(\omega \ell+\phi)}
\sum_{k=0}^{K_h-1}h_k s_m[\ell-k]+n[\ell],
\qquad \ell=1,\ldots,L,
\label{eq:received_signal}
\end{equation}
where $\{h_k\}_{k=0}^{K_h-1}$ denotes the channel response, $\omega$ and $\phi$ are frequency and phase offsets, and $n[\ell]$ is additive noise. This expression is a general abstraction, the exact channel and impairment distributions follow each benchmark protocol.

The receiver stores an observation as
\begin{equation}
\mathbf{x}
=\begin{bmatrix}
\operatorname{Re}(r[1{:}L])\\
\operatorname{Im}(r[1{:}L])
\end{bmatrix}
\in\mathbb{R}^{2\times L}.
\end{equation}
Given a labeled dataset $\mathcal{D}=\{(\mathbf{x}_n,m_n)\}_{n=1}^{N}$, AMR learns a posterior classifier $p_{\boldsymbol{\theta}}(m\mid\mathbf{x},L)$. Its primary objective is the empirical cross-entropy
\begin{equation}
\mathcal{L}_{\mathrm{cls}}
=-\frac{1}{N}\sum_{n=1}^{N}
\log p_{\boldsymbol{\theta}}(m_n\mid\mathbf{x}_n,L_n).
\label{eq:classification_loss}
\end{equation}
The proposed classifier separates a common representation template from configuration-specific expert processing. The short and long configurations retain the same backbone topology but use different expert inductive biases and integration interfaces.

\subsection{Unified backbone--expert organization}
\begin{figure}[t]
\centering
\includegraphics[width=\textwidth]{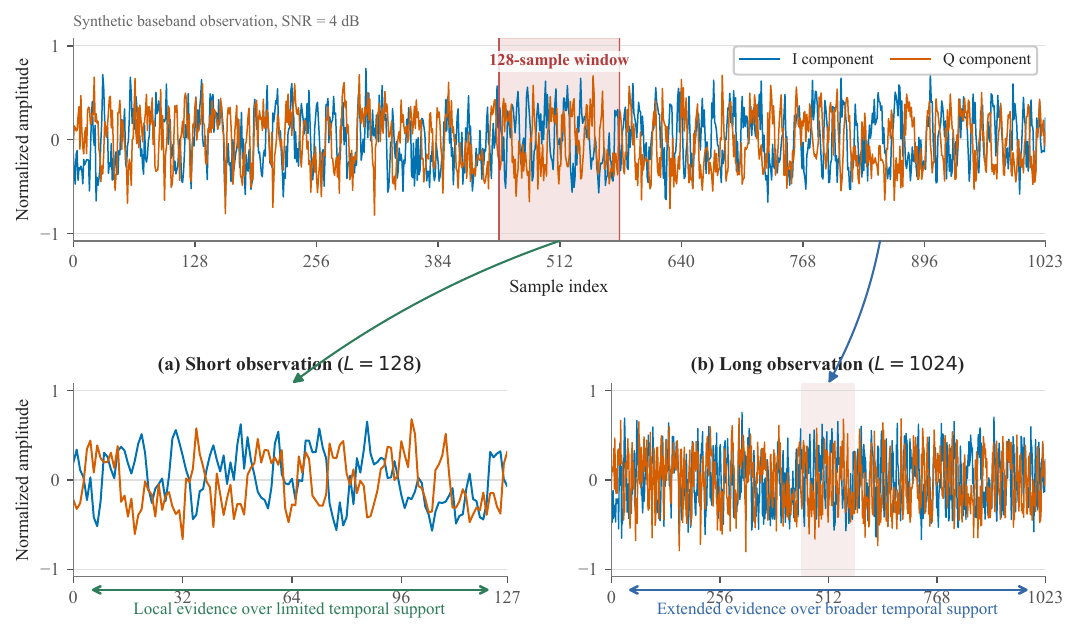}
\caption{Illustration of short and long observation budgets using the same synthetic QPSK signal realization. The 128-sample observation is a contiguous window of the 1024-sample observation. The shorter window contains signal evidence over limited temporal support, whereas the longer window retains evidence distributed over a broader temporal range.}
\label{fig:observation_budget}
\end{figure}

The observation budget changes the amount and temporal span of the evidence available to the classifier, while the signal and channel distribution determine how informative that evidence is. Figure~\ref{fig:observation_budget} illustrates the temporal-support difference using two nested windows from the same synthetic QPSK signal realization. The 128-sample window provides fewer samples from which to extract discriminative structure, motivating investigation of explicit local lag-dependent relations before contextual sequence modeling. The 1024-sample observation covers a broader temporal range, motivating investigation of multi-scale refinement and complementary classifiers. Accordingly, the two evaluated configurations retain a common backbone topology but employ different expert inductive biases and integration interfaces for their complete benchmark conditions.

To formalize this organization, let $f_{\mathrm{s}}(\cdot;\boldsymbol{\theta}_{\mathrm{s}})$ and $f_{\mathrm{l}}(\cdot;\boldsymbol{\theta}_{\mathrm{l}})$ denote the short- and long-observation configurations:
\begin{align}
\mathbf{h}_b
&=\mathcal{B}(\mathbf{x};\boldsymbol{\theta}^{\mathcal{B}}_b),\\
f_b(\mathbf{x};\boldsymbol{\theta}_b)
&=\mathcal{I}_b\!\left(
\mathbf{h}_b,
\mathcal{E}_b(\mathbf{x},\mathbf{h}_b;\boldsymbol{\theta}^{\mathcal{E}}_b);
\boldsymbol{\theta}^{\mathcal{I}}_b
\right),
\qquad b\in\{\mathrm{s},\mathrm{l}\},
\label{eq:regime_configuration}
\end{align}
where $\mathcal{B}$ denotes the common backbone topology, $\mathcal{E}_b$ is the configuration-specific expert, and $\mathcal{I}_b$ is its integration interface. The notation allows an expert to use the raw observation, an intermediate backbone representation, or both. The short interface injects position-aligned relation descriptors into the sequence tokens before contextual encoding. The long interface applies residual feature refinement and combines independently formed convolutional and state-space decisions.

\begin{figure}[H]
\centering
\includegraphics[width=\textwidth]{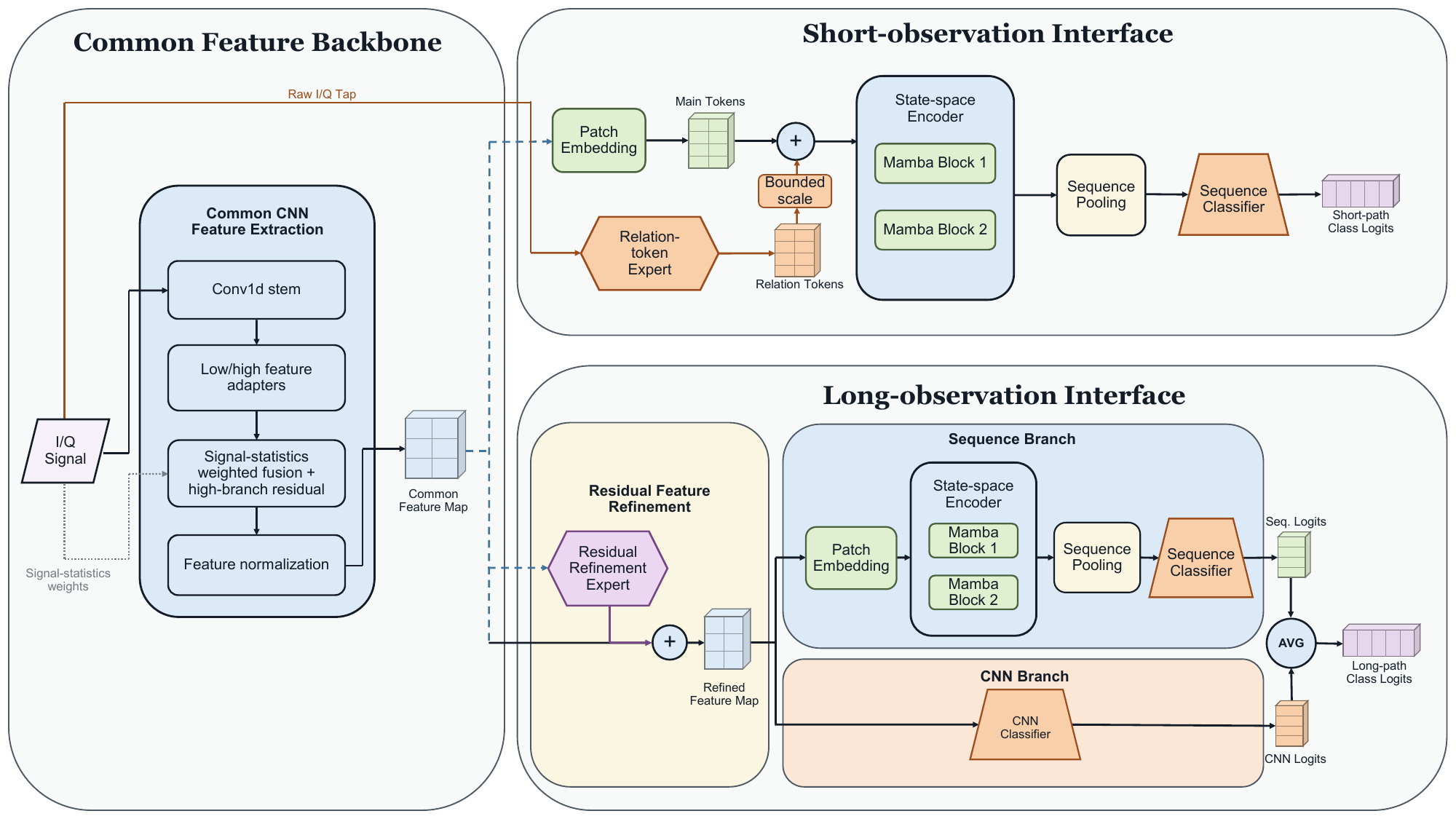}
\caption{Overall backbone--expert framework. The short and long configurations use the same feature-extraction and state-space topology but independently trained weights. They differ in expert inductive bias and integration interface.}
\label{fig:backbone_expert_framework}
\end{figure}

Each benchmark uses the corresponding configuration, which is trained and executed independently. The two configurations use the same backbone topology and feature dimensions but have different learned weights. In the experiments, $f_{\mathrm{s}}$ is instantiated for RML2016.10b at its native length of 128 and $f_{\mathrm{l}}$ for HisarMod2019 at its native length of 1024. The cross-configuration study reverses these assignments to assess whether the proposed expert designs contribute beyond a uniformly stronger configuration. The resulting organization is summarized in Fig.~\ref{fig:backbone_expert_framework}.

\subsection{Common convolutional state-space backbone}
The two configurations use a common architectural template for initial feature extraction and sequence encoding. The experiments use a parameter-matched ordinary Conv1d stem. The stem applies a depthwise-separable convolution~\cite{howard2017mobilenet} with a kernel size of 31, followed by channel recalibration and two convolutional mixing layers. This choice keeps the stem compact and focuses the architecture on the configuration-specific experts.

The stem output $\mathbf{F}_0$ is processed by two complementary adapters. A depthwise branch with kernel size 7 emphasizes local context, whereas a standard convolutional branch with kernel size 3 provides channel mixing. A signal-statistics mixer maps eight global statistics $\boldsymbol{\xi}(\mathbf{x})$ to two normalized weights $\pi_1$ and $\pi_2$. These statistics are the means and standard deviations of I and Q, the mean and standard deviation of amplitude, mean power, and the amplitude coefficient of variation. The common feature map is
\begin{equation}
\mathbf{F}
=\operatorname{BN}\!\left(
\pi_1 A_1(\mathbf{F}_0)
+\pi_2 A_2(\mathbf{F}_0)
+A_2(\mathbf{F}_0)
\right).
\label{eq:common_feature_map}
\end{equation}
The additional $A_2(\mathbf{F}_0)$ term forms a residual reference and prevents the mixed representation from completely suppressing the standard convolutional branch. This mixer is internal to the common backbone template; it does not select the short or long expert configuration.

A grouped patch embedding with kernel size 16 and stride 8 converts $\mathbf{F}$ into 96-dimensional tokens. Channel shuffle~\cite{zhang2018shufflenet} and a pointwise projection restore interaction across convolution groups. Context is then modeled by two residual Mamba blocks~\cite{gu2024mamba}:
\begin{equation}
\mathbf{X}^{(k+1)}
=\mathbf{X}^{(k)}
+\operatorname{Dropout}\!\left(
\operatorname{Mamba}_k(\operatorname{LN}(\mathbf{X}^{(k)}))
\right),
\qquad k=0,1.
\label{eq:mamba_backbone}
\end{equation}
Each block uses model width 96 and state dimension 24. The selective state-space operator provides content-dependent sequence propagation with linear scaling in token length. Mean and maximum pooling of the encoded tokens are concatenated and passed to a linear classifier. The short configuration inserts relation tokens before the Mamba encoder, whereas the long configuration refines $\mathbf{F}$ before patch embedding and adds a second classifier at the output interface.

\subsection{Short-sequence relation-aware token augmentation}
Complex-valued convolution provides an effective way to preserve the coupling between the I and Q components~\cite{xin2025cppcnet}. For the short configuration, we pursue a complementary design that makes selected local complex relations explicit before representation learning. This reduces the burden on a learned branch to recover these relations solely from stacked operations and provides direct control over the temporal offsets being modeled. The resulting Polar-Lag module constructs a compact set of amplitude and lag-relation descriptors, followed by a lightweight learned token projection.

To construct these descriptors, let $z_t=i_t+jq_t$ and define its amplitude and power as
\begin{equation}
a_t=\sqrt{i_t^2+q_t^2+\epsilon},
\qquad p_t=i_t^2+q_t^2,
\qquad \Delta a_t=a_t-a_{t-1}.
\end{equation}
For each lag $\ell$ in $\mathcal{S}=\{1,2,4,8,16\}$, we compute
\begin{align}
d_{t,\ell}
&=|z_t-z_{t-\ell}| 
=\sqrt{(i_t-i_{t-\ell})^2+(q_t-q_{t-\ell})^2+\epsilon},\\
c_{t,\ell}
&=z_tz_{t-\ell}^{*}
=r_{t,\ell}+j u_{t,\ell},\\
r_{t,\ell}
&=i_ti_{t-\ell}+q_tq_{t-\ell},\\
u_{t,\ell}
&=q_ti_{t-\ell}-i_tq_{t-\ell}.
\end{align}
Samples before the start of the observation are set to zero. The distance $d_{t,\ell}$ measures local displacement in the complex plane, while $r_{t,\ell}$ and $u_{t,\ell}$ retain the in-phase and quadrature components of lagged complex correlation. Under a common phase rotation $z'_t=z_te^{j\varphi}$, both $d_{t,\ell}$ and $c_{t,\ell}$ remain unchanged. This invariance applies to the relation channels, not to the complete network, which also receives raw I/Q samples.

The final short path concatenates
\begin{equation}
\mathbf{p}_t=
\operatorname{concat}
\left(
i_t,q_t,a_t,p_t,\Delta a_t,
\{d_{t,\ell},r_{t,\ell},u_{t,\ell}\}_{\ell\in\mathcal{S}}
\right).
\label{eq:polar_lag_features}
\end{equation}
The resulting feature vector contains 20 channels and retains relative phase information through the real and imaginary correlation components.

\begin{figure}[H]
\centering
\includegraphics[width=\textwidth]{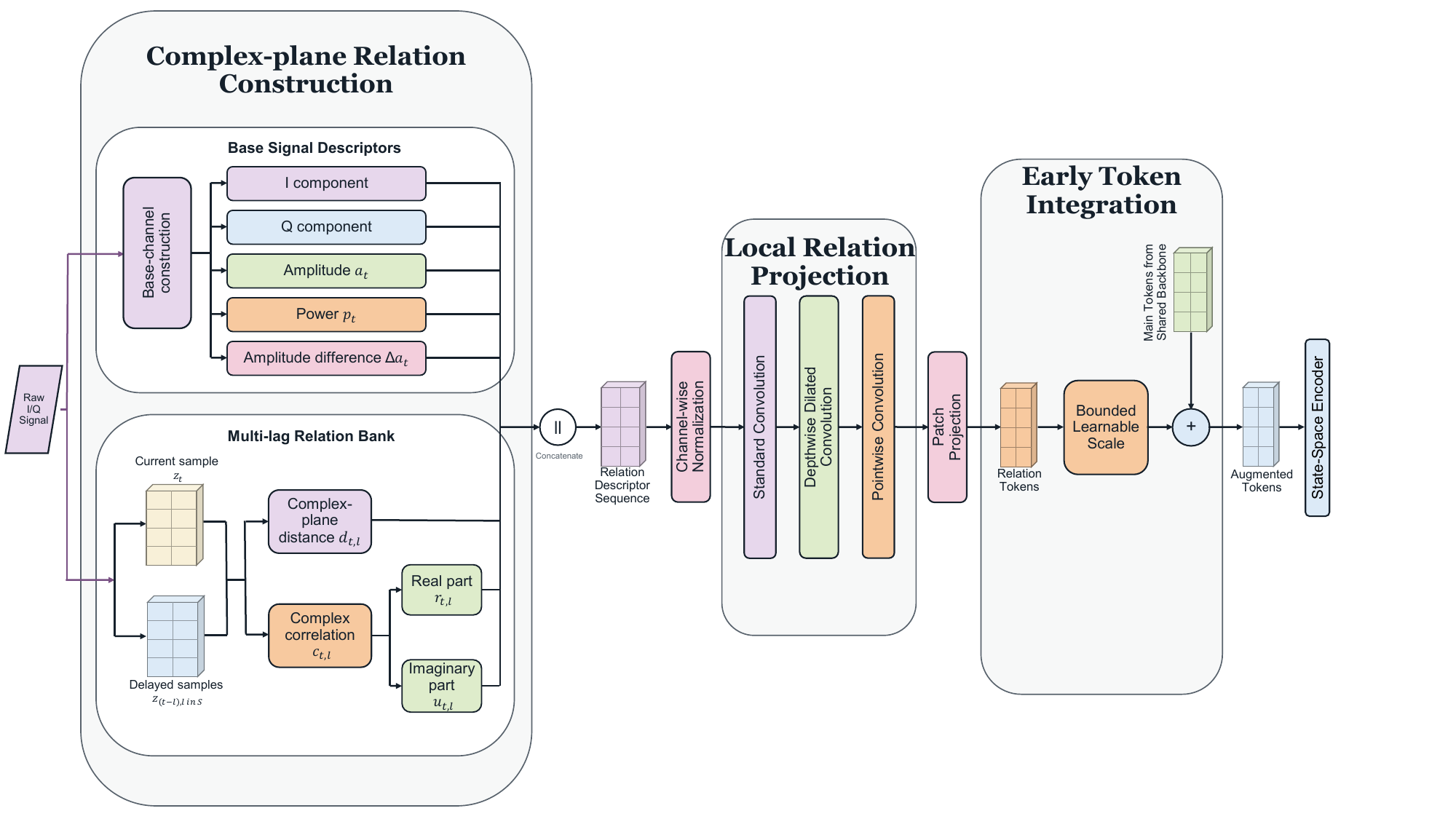}
\caption{Short-sequence relation-aware token augmentation. Base signal descriptors and multi-lag distance and complex-correlation relations are locally projected into relation tokens, multiplied by a bounded learned scale, and added to the main tokens before state-space encoding.}
\label{fig:short_relation_token_module}
\end{figure}

The features in Eq.~\eqref{eq:polar_lag_features} are first normalized channel-wise. We denote the resulting sequence by $\mathbf{P}$. A local projection $E_{\mathrm{rel}}$ then applies a standard convolution, a depthwise dilated convolution, and a pointwise convolution. The first operation mixes the heterogeneous relation channels over adjacent samples. The depthwise dilated operation enlarges the local temporal receptive field at limited computational cost~\cite{yu2016dilated}, and the pointwise operation recombines the resulting channels. A patch projection with kernel size 16 and stride 8 then maps the features to $d=96$ relation tokens:
\begin{equation}
\mathbf{T}_{\mathrm{rel}}=E_{\mathrm{rel}}(\mathbf{P}).
\end{equation}
In parallel, the common stem and patch embedding produce the main tokens $\mathbf{T}_{\mathrm{main}}$ with the same temporal resolution. The two streams are integrated before the sequence encoder:
\begin{equation}
\mathbf{T}_{\mathrm{s}}
=\mathbf{T}_{\mathrm{main}}
+\bar{\alpha}_{\mathrm{s}}\mathbf{T}_{\mathrm{rel}},
\qquad
\bar{\alpha}_{\mathrm{s}}
=\operatorname{clip}(\alpha_{\mathrm{s}},0,\alpha_{\max}),
\label{eq:short_token_fusion}
\end{equation}
where $\alpha_{\mathrm{s}}$ is a learned scalar and $\alpha_{\max}=0.25$. The augmented tokens are processed by the short-path state-space sequence encoder and pooling classifier. Integrating at the token interface allows the contextual encoder to model interactions between the backbone and relation tokens, whereas logit-level integration combines only their final decisions. The complete relation-token interface is illustrated in Fig.~\ref{fig:short_relation_token_module}.

Ablations with the same three seeds show that removing the relation branch or its distance channels mainly degrades low-SNR recognition, while high-SNR accuracy changes little. The early token interface also performs consistently better than the corresponding late logit interface. These observations motivate the use of the relation tokenizer and early integration in the final short path.

\subsection{Long-sequence residual refinement and classifier collaboration}
Long observations provide more temporal evidence, but channel impairments can spread unreliable responses over a wider feature sequence. One approach is to reconstruct a cleaner signal before classification with a denoising autoencoder~\cite{long2025dae}. Such reconstruction introduces a separate decoder and optimizes signal fidelity in addition to class discrimination. Related residual denoising work also motivates separating a corrective residual from the main task mapping~\cite{lin2026drdd}. Following this general principle, the proposed long configuration performs residual refinement directly in the discriminative feature space. It does not reconstruct the waveform or execute a diffusion process.

Let $\mathbf{F}\in\mathbb{R}^{C_f\times L}$ denote the common feature map in Eq.~\eqref{eq:common_feature_map}. The residual expert first forms a gated value representation
\begin{equation}
\mathbf{V}
=\psi_{v}(\mathbf{F})
\odot\sigma\!\left(\psi_{i}(\mathbf{F})\right),
\label{eq:long_gated_value}
\end{equation}
where $\psi_v$ is a pointwise projection followed by normalization and activation, $\psi_i$ is a pointwise input gate, $\sigma$ denotes the sigmoid function, and $\odot$ is element-wise multiplication. Three depthwise temporal branches process $\mathbf{V}$ with kernel--dilation pairs $(15,1)$, $(31,2)$, and $(63,2)$. Their effective receptive fields are 15, 61, and 125 samples, respectively. The branch outputs are concatenated and fused by a pointwise projection:
\begin{equation}
\mathbf{H}
=\psi_f\!\left(
\operatorname{concat}
\left[D_{15,1}(\mathbf{V}),D_{31,2}(\mathbf{V}),D_{63,2}(\mathbf{V})\right]
\right).
\label{eq:long_multiscale}
\end{equation}
The depthwise branches capture complementary temporal extents without the cost of dense large-kernel convolution, while $\psi_f$ restores cross-channel interaction. An output gate and a learned residual scale produce
\begin{equation}
R_{\mathrm{l}}(\mathbf{F})
=\beta_{\mathrm{l}}
\,\sigma\!\left(\psi_o(\mathbf{F})\right)
\odot\mathbf{H},
\label{eq:long_residual}
\end{equation}
where $\beta_{\mathrm{l}}$ is initialized to 0.15. The expert therefore predicts a gated correction to the shared representation rather than a replacement feature map.

The proposed expert uses a sample-dependent residual gate $g_{\mathrm{l}}(\mathbf{F})$ obtained from global average pooling and a small multilayer perceptron. A fixed upper coefficient $\alpha_{\mathrm{l}}$ controls the maximum contribution of the expert. The refined feature map is
\begin{equation}
\widetilde{\mathbf{F}}
=\mathbf{F}
+\alpha_{\mathrm{l}}
\,g_{\mathrm{l}}(\mathbf{F})
\,R_{\mathrm{l}}(\mathbf{F}),
\label{eq:long_refinement}
\end{equation}
where $\alpha_{\mathrm{l}}=0.85$ in the reported experiments, so the sample-dependent path weight $\alpha_{\mathrm{l}}g_{\mathrm{l}}(\mathbf{F})$ is bounded by 0.85.

\begin{figure}[H]
\centering
\includegraphics[width=\textwidth]{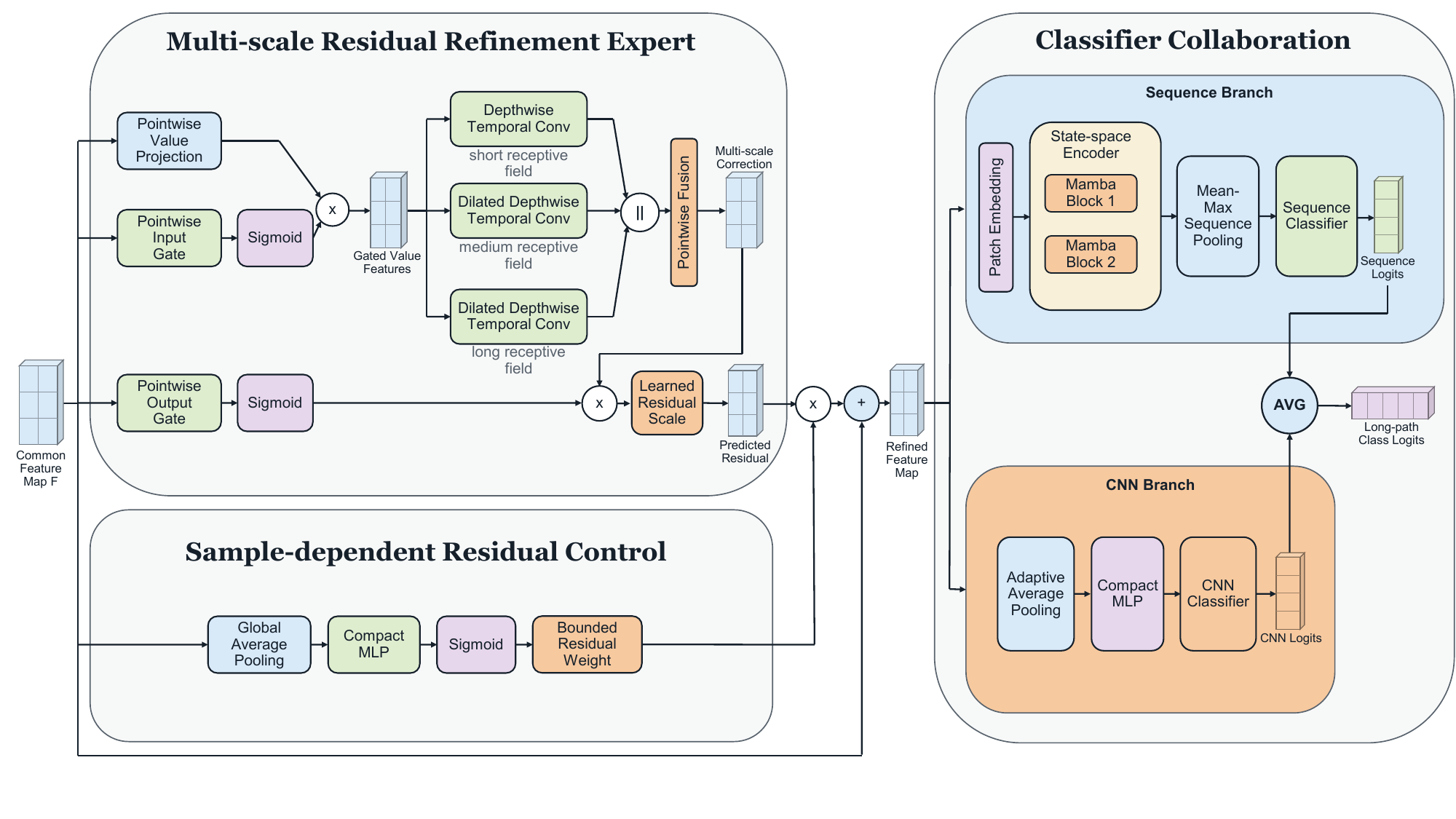}
\caption{Long-sequence residual--classifier organization. Multi-scale gated residual refinement corrects the common feature map before patch embedding. The state-space and CNN branches then form separate logits that are combined by fixed averaging.}
\label{fig:long_residual_classifier_module}
\end{figure}

Two classifiers provide complementary views of $\widetilde{\mathbf{F}}$. The sequence path converts it into patch tokens, applies the state-space encoder, and pools the encoded sequence to obtain logits $\mathbf{z}_{\mathrm{seq}}$. In parallel, the convolutional classification expert adaptively pools the feature map to 16 temporal positions and applies a compact multilayer perceptron to obtain $\mathbf{z}_{\mathrm{cnn}}$. Their decisions are combined by fixed-average logit fusion:
\begin{equation}
\mathbf{z}
=\frac{1}{2}\left(\mathbf{z}_{\mathrm{cnn}}+\mathbf{z}_{\mathrm{seq}}\right).
\label{eq:long_classifier_fusion}
\end{equation}
The sequence classifier models ordered contextual dependencies, whereas the CNN expert provides a direct classification view of the refined feature map. Averaging retains their separately formed class evidence without adding a sample-dependent fusion network. This is a logit-level collaboration interface, in contrast to the short configuration's token-level relation interface. Feature concatenation and learned logit gating are evaluated as alternative interfaces in Section~\ref{sec:results}. The residual--classifier organization is shown in Fig.~\ref{fig:long_residual_classifier_module}.

The residual and classification experts therefore act at different stages: the former corrects the feature sequence before contextual encoding, while the latter introduces a complementary decision after encoding. Their contributions and the choice of integration interface are examined separately in the experiments.

\subsection{Training objectives}
Both configurations are optimized only with the standard cross-entropy classification loss in Eq.~\eqref{eq:classification_loss}. All ablations use the same primary supervision.

\subsection{Computational complexity}
The two configurations are trained and executed separately. We therefore report active-configuration parameters and operations for the short and long configurations separately. The present experiments do not execute both configurations for each input. Complexity is expressed in multiply--accumulate operations (MACs). Convolutional and linear layers are counted from their realized tensor shapes. For each Mamba block, we additionally count the input, state, time-step, output projections, the causal depthwise convolution and the real-valued selective scan. For batch size $B$, token length $T$, inner width $d_{\mathrm{i}}$, and state size $d_{\mathrm{s}}$, the scan contribution is estimated as
\begin{equation}
\mathcal{C}_{\mathrm{scan}} = 9BTd_{\mathrm{i}}d_{\mathrm{s}} + 2BTd_{\mathrm{i}},
\label{eq:selective_scan_complexity}
\end{equation}
where the final term accounts for the skip and output-gating operations. One MAC corresponds approximately to two scalar floating-point operations. This accounting avoids omitting the fused selective-scan kernel from the end-to-end estimate.

\section{Experimental setup}
\label{sec:experiments}

\subsection{Datasets and protocols}
\begin{table}[H]
\centering
\caption{Dataset and evaluation protocols.}
\label{tab:datasets}
\small
\resizebox{\textwidth}{!}{%
\begin{tabular}{lcccc}
\toprule
Dataset & Length & Classes & SNR range & Split (train:validation:test) \\
\midrule
RML2016.10b & 128 & 10 & $-20{:}2{:}18$ dB & 6:2:2 \\
HisarMod2019 & 1024 & 26 & $-20{:}2{:}18$ dB & 8:2:5 \\
\bottomrule
\end{tabular}%
}
\end{table}

The short-observation and long-observation configurations are evaluated on RML2016.10b~\cite{oshea2016dataset} and HisarMod2019~\cite{tekbiyik2020hisar}, respectively. The RML2016.10b files used in this study were obtained from a public Kaggle mirror, while HisarMod2019 was obtained from IEEE DataPort. RML2016.10b contains simulated I/Q observations with 10 modulation classes, whereas HisarMod2019 contains 26 classes generated under multiple fading conditions. We retain the native observation length and the split protocol reported in Table~\ref{tab:datasets} for each formal configuration.

\subsection{Compared methods}
We compare against representative AMR methods using graph, complex-valued, multimodal, multi-scale, attention, denoising, and language-model representations. STF-GCN constructs adaptive spatial--temporal--frequency graphs~\cite{shao2025stfgcn}. IQFormer and MCANet use multimodal feature collaboration~\cite{shao2025iqformer,jiang2025mcanet}, while CPPCNet uses lightweight complex-valued partial pointwise convolution~\cite{xin2025cppcnet}. Recent HisarMod2019 comparisons include MAWDN, DFENet, WACN, and MILAFormer~\cite{qin2025mawdn,le2026dfenet,feng2025wacn,zhao2026milaformer}. BioLAMR adapts a pretrained language model with dual-domain signal features~\cite{mao2026biolamr}. DAE-CNN-BiLSTM combines denoising reconstruction with convolutional and recurrent classifiers~\cite{long2025dae}.

The reported studies use different data partitions and model-selection protocols. Table~\ref{tab:reported_comparison} records the stated protocol for each dataset so that the literature results can be interpreted in context.

\subsection{Implementation details}
All models are implemented in PyTorch with the CUDA Mamba implementation. Experiments are conducted on an NVIDIA RTX 4070 SUPER GPU with an Intel Core i5-13600KF CPU. Both configurations use the parameter-matched convolutional state-space backbone with model width 96, two Mamba blocks of state dimension 24, and grouped patch embedding with kernel size 16 and stride 8. The short configuration adds the distance and complex-correlation relation channels, whereas the long configuration uses residual refinement and averages the CNN and sequence logits.

For RML2016.10b, all configurations are trained from random initialization with AdamW, learning rate $2.5\times10^{-4}$, weight decay $10^{-4}$, batch size 800, and at most 80 epochs. All short experiments use the same augmentation protocol: circular shifts of at most four samples, multiplicative gain perturbation with standard deviation 0.05, phase rotation bounded by 0.35 radians, and same-class segment substitution with probability 0.35 and segment length 16. The checkpoint with highest validation accuracy is retained with early-stopping patience 14. For HisarMod2019, models are trained from random initialization with Adam, learning rate $10^{-3}$, weight decay $10^{-4}$, batch size 400, and at most 200 epochs. The learning rate is halved after five validation-loss plateaus, early stopping uses patience 10, and the checkpoint is selected by validation accuracy averaged over samples with $\mathrm{SNR}\leq0$.

The RML2016.10b experiments use training seeds 2051, 2052, and 2053. The HisarMod2019 experiments use seeds 2028, 2029, and 2030 with data split seed 2024. For HisarMod2019, 20\% of the official training partition is reserved for validation, giving the 8:2:5 ratio in Table~\ref{tab:datasets}; the official test partition remains untouched during model selection.

\subsection{Evaluation metrics and statistical analysis}
We report overall average accuracy (OAA), mean accuracy for $\mathrm{SNR}<0$, and mean accuracy for $\mathrm{SNR}\geq0$. We additionally report $\mathrm{SNR}\leq-10$ as a descriptive very-low-SNR indicator. Results are summarized as the mean $\pm$ sample standard deviation over three independent training seeds, providing a reproducibility-oriented estimate of run-to-run variation. Same-seed OAA differences are used as descriptive stability checks rather than as formal population-level significance tests.

\subsection{Native-length cross-configuration protocol}
To verify that the observed gains are associated with the proposed expert designs rather than one configuration being uniformly stronger, we construct a $2\times2$ native-length cross-configuration evaluation. On each dataset, both the short-observation and long-observation configurations are evaluated while retaining the native input length and original test protocol, so no signal samples are discarded. The configuration used for that dataset is compared with the cross-applied alternative. This comparison evaluates the suitability of each expert design under the two benchmark conditions. Because dataset identity and observation length remain coupled, it does not isolate a causal effect of length.

\subsection{Controlled observation-window protocol}
To partially separate temporal support from dataset identity, we conduct an additional study within HisarMod2019. Each native 1024-sample observation is represented by nested center windows of length 128, 256, and 512, together with the unmodified 1024-sample observation. No padding, resampling, or cross-record concatenation is used. Sample identity, modulation label, SNR, official train/test membership, and the train/validation split are therefore unchanged across lengths. At each length, the short and long configurations are trained independently from random initialization using seeds 2028--2030 and the HisarMod2019 optimization protocol in Section~\ref{sec:experiments}. The study measures how each configuration uses increasing temporal support within one data source.

\section{Results}
\label{sec:results}

\subsection{Comparison with literature-reported methods}

Table~\ref{tab:reported_comparison} places the proposed configurations alongside representative recent results. Values for competing methods are taken from their original publications, whereas the proposed results are the mean and sample standard deviation over the three seeds defined in Section~\ref{sec:experiments}.

\begin{table}[H]
\centering
\caption{Reported OAA (\%), proposed SNR-stratified results, and dataset-specific protocols. Ratios use train:validation:test order unless marked as train/validation or train/test; a dash denotes an unreported result.}
\label{tab:reported_comparison}
\resizebox{\textwidth}{!}{%
\begin{tabular}{lcccc}
\toprule
Method & RML2016.10b OAA & HisarMod2019 OAA & RML2016.10b protocol & HisarMod2019 protocol \\
\midrule
STF-GCN~\cite{shao2025stfgcn} & 66.04 & -- & 6:2:2 & -- \\
IQFormer~\cite{shao2025iqformer} & 65.65 & 76.32 & 6:2:2 & 8:2:5 \\
CPPCNet~\cite{xin2025cppcnet} & 66.38 & 83.50 & 8:2 train/val & 8:2 train/val \\
BioLAMR~\cite{mao2026biolamr} & $67.43\pm0.27$ & -- & 8:1:1 & -- \\
MCANet~\cite{jiang2025mcanet} & 66.53 & 76.58 & 7:2:1 & 7:2:1 \\
MAWDN~\cite{qin2025mawdn} & -- & 74.40 & -- & 8:2:5 \\
MILAFormer~\cite{zhao2026milaformer} & -- & 77.45 & -- & 8:2:5 \\
DFENet~\cite{le2026dfenet} & -- & 82.76 & -- & 2:1 train/test \\
WACN~\cite{feng2025wacn} & 64.70 & 95.54 & Not stated & Not stated \\
DAE-CNN-BiLSTM~\cite{long2025dae} & -- & 86.93 & -- & 8:2:5 \\
Proposed & $67.28\pm0.14$ & $87.19\pm0.77$ & 6:2:2 & 8:2:5 \\
\midrule
\multicolumn{5}{l}{\textit{Proposed configurations: SNR-stratified results}} \\
Configuration & OAA & $\mathrm{SNR}\leq-10$ & $\mathrm{SNR}<0$ & $\mathrm{SNR}\geq0$ \\
RML2016.10b, short & $67.279\pm0.135$ & $23.725\pm0.602$ & $41.121\pm0.267$ & $93.436\pm0.033$ \\
HisarMod2019, long & $87.186\pm0.773$ & $71.126\pm1.710$ & $75.577\pm1.484$ & $98.794\pm0.062$ \\
\bottomrule
\end{tabular}
}
\end{table}

Under the explicitly reported 6:2:2 protocol on RML2016.10b, the proposed short configuration achieves the best OAA among the comparable entries listed in Table~\ref{tab:reported_comparison}. Its result is also within 0.15 percentage points of the highest reported BioLAMR result, although BioLAMR uses a different data partition. On HisarMod2019, the proposed long configuration achieves the highest OAA among the listed methods that explicitly report the same 8:2:5 protocol, exceeding DAE-CNN-BiLSTM by 0.26 percentage points. These comparisons indicate competitive state-of-the-art performance under protocol-matched settings, rather than a protocol-independent ranking across all published results. The lower panel of Table~\ref{tab:reported_comparison} gives the proposed SNR-stratified results.

\subsection{Short-sequence ablation}

Table~\ref{tab:short_ablation} evaluates the relation descriptors and their integration interface on RML2016.10b. Setting the relation gate to zero removes the complete relation-token update while preserving the backbone and the main sequence classifier. It is therefore a control for the contribution of the relation branch as a whole, rather than an ablation of distance channels alone. The late-fusion variant processes the same relation descriptors with a separate classification head and adds its gated logits after sequence classification, so it tests the integration location while retaining the relation information.

\begin{table}[H]
\centering
\caption{Short-configuration ablation on RML2016.10b (\%, mean $\pm$ sample standard deviation over seeds 2051--2053).}
\label{tab:short_ablation}
\scriptsize
\resizebox{\textwidth}{!}{%
\begin{tabular}{lccccc}
\toprule
Variant & OAA & $\mathrm{SNR}\leq-10$ & $\mathrm{SNR}<0$ & $\mathrm{SNR}\geq0$ & $\Delta$OAA \\
\midrule
Full short-observation configuration & $67.279\pm0.135$ & $23.725\pm0.602$ & $41.121\pm0.267$ & $93.436\pm0.033$ & -- \\
w/o correlation features & $66.859\pm1.180$ & $21.944\pm3.164$ & $40.267\pm2.330$ & $93.450\pm0.032$ & $-0.420$ \\
w/o distance features & $65.224\pm0.245$ & $17.526\pm0.067$ & $36.968\pm0.452$ & $93.480\pm0.044$ & $-2.054$ \\
Relation gate $=0$ & $65.250\pm0.223$ & $17.398\pm0.069$ & $37.054\pm0.388$ & $93.446\pm0.071$ & $-2.029$ \\
Late logit fusion & $65.749\pm0.163$ & $20.089\pm0.202$ & $38.304\pm0.333$ & $93.193\pm0.060$ & $-1.530$ \\
\bottomrule
\end{tabular}%
}
\end{table}

Relative to the zero-gate variant, the full configuration improves OAA by 2.029 percentage points and $\mathrm{SNR}<0$ accuracy by 4.067 points, while the $\mathrm{SNR}\geq0$ accuracies differ by only 0.010 points. Removing distance features produces a similar OAA loss of 2.054 points. The nearly identical losses of the w/o-distance and zero-gate controls indicate that distance accounts for most of the measured relation-branch gain. Correlation features produce a modest mean difference relative to run-to-run variation; they are therefore retained as a complementary relation descriptor without attributing an independently validated accuracy gain to this component. Early token augmentation exceeds late logit fusion by 1.530 points, with most of the difference again occurring below 0 dB. This difference is consistent with the two interfaces exposing the relation information at different stages: early addition lets the state-space encoder contextualize relation and backbone tokens jointly, whereas late fusion can only combine the separately formed final decisions.

\subsection{Long-sequence expert and interface analysis}

Table~\ref{tab:long_ablation} evaluates the long-configuration classifier interface and expert contributions on HisarMod2019. All variants retain the convolutional state-space backbone and the same classification objective. Removing residual refinement leaves the two classifier branches and their decision interface unchanged. The learned-gate variant uses a feature-dependent coefficient instead of equal logit weights, while feature concatenation combines the sequence and convolutional features before a classifier. CNN-only retains the convolutional classifier, sequence-only retains the sequence classifier, and the both-experts-off control removes both the residual and CNN experts.

\begin{table}[H]
\centering
\caption{Long-configuration expert and interface analysis on HisarMod2019 (\%, mean $\pm$ sample standard deviation over seeds 2028--2030).}
\label{tab:long_ablation}
\scriptsize
\resizebox{\textwidth}{!}{%
\begin{tabular}{lccccc}
\toprule
Variant & OAA & $\mathrm{SNR}\leq-10$ & $\mathrm{SNR}<0$ & $\mathrm{SNR}\geq0$ & $\Delta$OAA \\
\midrule
Equal-weight logit averaging (formal) & $87.186\pm0.773$ & $71.126\pm1.710$ & $75.577\pm1.484$ & $98.794\pm0.062$ & -- \\
w/o residual refinement & $79.893\pm0.365$ & $55.963\pm0.569$ & $61.924\pm0.547$ & $97.862\pm0.197$ & $-7.293$ \\
Learned logit gate & $86.963\pm0.622$ & $70.640\pm1.316$ & $75.131\pm1.213$ & $98.795\pm0.076$ & $-0.223$ \\
Feature concatenation & $85.186\pm0.258$ & $67.209\pm0.686$ & $71.932\pm0.523$ & $98.440\pm0.017$ & $-2.000$ \\
CNN classifier only & $86.013\pm0.319$ & $68.801\pm0.811$ & $73.410\pm0.670$ & $98.617\pm0.039$ & $-1.173$ \\
Sequence classifier only & $76.566\pm0.119$ & $56.938\pm0.040$ & $60.216\pm0.039$ & $92.915\pm0.215$ & $-10.620$ \\
Residual and CNN experts off & $72.770\pm0.742$ & $53.410\pm0.719$ & $56.103\pm0.660$ & $89.436\pm0.825$ & $-14.416$ \\
\bottomrule
\end{tabular}%
}
\end{table}

Removing residual refinement while retaining equal-weight decision collaboration lowers OAA by 7.293 points and $\mathrm{SNR}<0$ accuracy by 13.653 points, confirming the importance of residual refinement. Equal-weight averaging and learned gating achieve comparable accuracy, while both outperform feature concatenation and the separately optimized single-branch controls. Since learned gating provides no repeatable gain while introducing an additional sample-dependent fusion module, fixed averaging is adopted as the simpler decision-level interface. Feature concatenation is 2.000 points lower, which favors collaboration between separately formed decisions over joint feature compression. The CNN classifier is stronger than the sequence classifier individually; however, adding the sequence decision to the CNN branch improves OAA by 1.173 points and $\mathrm{SNR}<0$ accuracy by 2.167 points. Sequence-only and both-experts-off variants produce larger losses.

\begin{figure}[H]
\centering
\includegraphics[width=\textwidth]{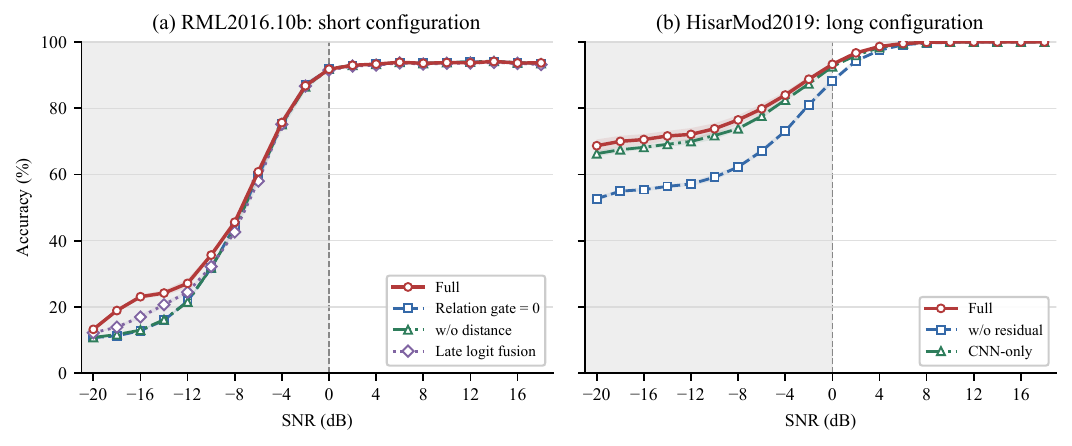}
\caption{Per-SNR accuracy of representative short- and long-configuration ablations. Curves show the mean over three seeds, shaded bands denote one sample standard deviation, and the gray background marks negative SNRs.}
\label{fig:snr_ablation_curves}
\end{figure}

Figure~\ref{fig:snr_ablation_curves} resolves the aggregate ablation results by SNR. For the short configuration, the gains from relation-aware token augmentation and its distance descriptor occur predominantly below 0 dB, while the curves nearly converge at nonnegative SNRs. For the long configuration, residual refinement yields its largest advantage in the negative-SNR region, and collaboration with the sequence classifier provides a smaller gain with the same concentration. The improvements are therefore associated mainly with difficult noise conditions rather than a uniform upward shift across SNRs. For clarity, Fig.~\ref{fig:snr_ablation_curves} visualizes only mechanism-representative variants; the complete set of ablations is reported in Tables~\ref{tab:short_ablation} and~\ref{tab:long_ablation}.

\subsection{Long-configuration mechanism diagnostics}

We further inspect the three selected formal checkpoints without retraining. For the residual interface, Table~\ref{tab:residual_gate_diagnostic} reports the bounded coefficient $\alpha_{\mathrm{l}}g_{\mathrm{l}}(\mathbf{F})$, its within-run variation across test samples, and the relative correction magnitude $\lVert\alpha_{\mathrm{l}}g_{\mathrm{l}}(\mathbf{F})\mathbf{R}\rVert_2/\lVert\mathbf{F}\rVert_2$. Each entry is the mean $\pm$ sample standard deviation across the three seed-level statistics.

\begin{table}[H]
\centering
\caption{Residual-interface diagnostics on the HisarMod2019 official test set. The coefficient is bounded above by 0.85. ``Within-run SD'' measures sample-to-sample coefficient variation within each checkpoint.}
\label{tab:residual_gate_diagnostic}
\small
\resizebox{0.94\textwidth}{!}{%
\begin{tabular}{lccc}
\toprule
Region & Residual coefficient & Within-run SD & Relative correction norm \\
\midrule
All SNRs & $0.642\pm0.033$ & $0.157\pm0.017$ & $1.194\pm0.096$ \\
$\mathrm{SNR}<0$ & $0.678\pm0.030$ & $0.132\pm0.020$ & $1.283\pm0.094$ \\
$\mathrm{SNR}\geq0$ & $0.606\pm0.035$ & $0.171\pm0.017$ & $1.106\pm0.097$ \\
\bottomrule
\end{tabular}%
}
\end{table}

The residual coefficient is larger on average below 0 dB, and the relative correction norm changes in the same direction. The nonzero within-run deviations show that the gate is not a fixed global multiplier. These statistics support sample-dependent residual adjustment and a stronger average intervention in the difficult SNR region. They do not imply that the gate is an explicit SNR estimator, because its input is the learned feature map and the coefficient distributions overlap across the two regions.

Table~\ref{tab:classifier_complementarity} examines the two logit branches inside the same jointly trained formal checkpoint. ``Fusion-only correct'' denotes samples for which the averaged logits predict the correct class although neither branch is individually top-1 correct. The margin correlation is the Pearson correlation between the two branches' true-class margins, where a margin is the true-class logit minus the largest competing logit.

\begin{table}[H]
\centering
\caption{Score-level complementarity of the jointly trained sequence and CNN classifiers on HisarMod2019 (\%, except correlation; mean $\pm$ sample standard deviation over seeds 2028--2030).}
\label{tab:classifier_complementarity}
\scriptsize
\resizebox{\textwidth}{!}{%
\begin{tabular}{lcccccc}
\toprule
Region & Sequence top-1 & CNN top-1 & Fused top-1 & Disagreement & Fusion-only correct & Margin correlation \\
\midrule
All SNRs & $60.537\pm0.718$ & $38.590\pm1.370$ & $87.186\pm0.773$ & $84.238\pm0.537$ & $6.524\pm0.287$ & $-0.439\pm0.015$ \\
$\mathrm{SNR}<0$ & $55.175\pm0.330$ & $24.382\pm0.886$ & $75.577\pm1.484$ & $91.468\pm0.103$ & $9.564\pm0.427$ & $-0.324\pm0.021$ \\
$\mathrm{SNR}\geq0$ & $65.900\pm1.360$ & $52.798\pm1.858$ & $98.794\pm0.062$ & $77.007\pm0.973$ & $3.485\pm0.189$ & $-0.676\pm0.004$ \\
\bottomrule
\end{tabular}%
}
\end{table}

The high prediction disagreement and negative margin correlation show that the two branches form different score residuals. Averaging can therefore recover decisions even when both separate argmax outputs are wrong, with the largest fusion-only recovery rate occurring below 0 dB. The extracted branch accuracies in Table~\ref{tab:classifier_complementarity} should not be equated with the separately optimized CNN-only and sequence-only controls in Table~\ref{tab:long_ablation}: the formal model applies supervision only to the averaged logits, so its two branch scores are free to co-adapt. Together, the retrained controls and the within-checkpoint diagnostic support decision-level collaboration while avoiding a claim that either jointly trained branch is a calibrated standalone classifier.

\subsection{Native-length cross-configuration comparison}
\begin{table}[H]
\centering
\caption{Native and cross-applied configurations at the original observation length of each dataset (OAA in \%, mean $\pm$ sample standard deviation). Intervals are computed from three same-seed OAA differences.}
\label{tab:cross_configuration}
\small
\resizebox{\textwidth}{!}{%
\begin{tabular}{lcccc}
\toprule
Dataset & Native configuration & Cross-applied configuration & Native advantage & Exploratory 95\% interval \\
\midrule
RML2016.10b ($L=128$) & Short: $67.279\pm0.135$ & Long: $65.326\pm0.067$ & $+1.952$ & $[1.726,2.179]$ \\
HisarMod2019 ($L=1024$) & Long: $87.186\pm0.773$ & Short: $67.023\pm1.393$ & $+20.163$ & $[16.074,24.252]$ \\
\bottomrule
\end{tabular}%
}
\end{table}

\begin{figure}[H]
\centering
\includegraphics[width=0.92\textwidth]{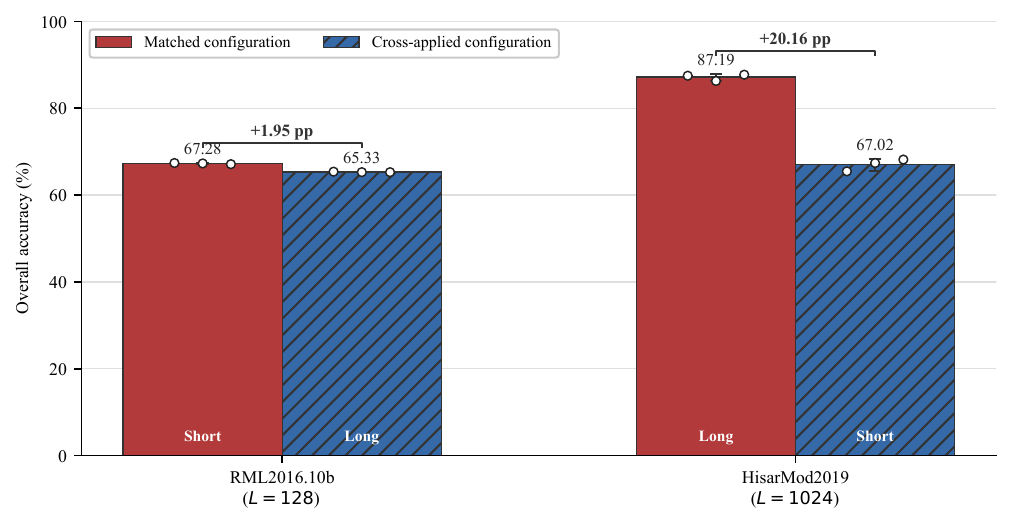}
\caption{Native and cross-applied configuration performance at the original observation length of RML2016.10b and HisarMod2019. Bars and error bars denote mean OAA and one sample standard deviation over three seeds, respectively; markers indicate individual seeds. Numbers above the brackets report the native-configuration advantage in percentage points.}
\label{fig:native_length_cross_configuration}
\end{figure}

As summarized in Table~\ref{tab:cross_configuration} and visualized in Fig.~\ref{fig:native_length_cross_configuration}, the native configuration wins in all six same-seed comparisons. On RML2016.10b, the short-observation configuration has a mean advantage of 1.952 points over the cross-applied long configuration. On HisarMod2019, the long-observation configuration has a larger mean advantage of 20.163 points over the cross-applied short configuration. These results show that neither configuration is uniformly stronger and support the use of different expert designs under the two evaluated benchmark conditions. Dataset identity and observation length nevertheless remain coupled in this comparison.

\subsection{Controlled observation-window comparison}

Table~\ref{tab:controlled_length} compares the two configurations using nested windows from the same HisarMod2019 observations. Long--Short differences are computed seed by seed before aggregation.

\begin{table}[H]
\centering
\caption{Controlled observation-window comparison on HisarMod2019 (OAA in \%, mean $\pm$ sample standard deviation over seeds 2028--2030).}
\label{tab:controlled_length}
\small
\begin{tabular}{lccc}
\toprule
Length & Short & Long & Long--Short \\
\midrule
128  & $61.783\pm0.952$ & $71.527\pm0.233$ & $9.744\pm0.719$ \\
256  & $65.090\pm0.646$ & $76.970\pm0.228$ & $11.881\pm0.482$ \\
512  & $67.433\pm0.310$ & $81.679\pm0.527$ & $14.247\pm0.225$ \\
1024 & $67.023\pm1.393$ & $87.186\pm0.773$ & $20.163\pm1.646$ \\
\bottomrule
\end{tabular}
\end{table}

The long configuration improves by 15.659 points from 128 to 1024 samples, whereas the short configuration improves by 5.240 points and saturates after 512 samples. More importantly, the Long--Short advantage increases monotonically with window length for every seed. This same-source result shows that the residual--classifier configuration uses additional temporal support more effectively on HisarMod2019. At the same nominal length of 128 samples, the preferred configuration differs between HisarMod2019 and RML2016.10b, indicating that temporal support interacts with the modulation and channel distributions of the data source when determining the effective expert design.

\subsection{Paired stability analysis}

Table~\ref{tab:paired_stability} reports the same-seed OAA differences between each formal configuration and its ablated counterpart. Positive values indicate an advantage for the formal configuration. The short-path columns correspond to seeds 2051--2053, and the long-path columns correspond to seeds 2028--2030. This table exposes the direction and magnitude of run-to-run differences directly; it is used as a reproducibility-oriented stability check rather than as a formal significance test.

\begin{table}[H]
\centering
\caption{Same-seed paired OAA differences (percentage points) for the short and long configurations. Positive values favor the corresponding formal configuration.}
\label{tab:paired_stability}
\scriptsize
\resizebox{\textwidth}{!}{%
\begin{tabular}{lrrrr}
\toprule
Comparison & Seed 1 & Seed 2 & Seed 3 & Mean $\Delta$OAA \\
\midrule
\multicolumn{5}{l}{\textit{Short configuration (seeds 2051, 2052, 2053)}} \\
Full $-$ w/o correlation & $-0.135$ & $-0.242$ & $+1.637$ & $+0.420$ \\
Full $-$ w/o distance & $+2.300$ & $+1.795$ & $+2.067$ & $+2.054$ \\
Full $-$ relation gate $=0$ & $+2.292$ & $+1.795$ & $+1.999$ & $+2.029$ \\
Full $-$ late fusion & $+1.487$ & $+1.557$ & $+1.546$ & $+1.530$ \\
\midrule
\multicolumn{5}{l}{\textit{Long configuration (seeds 2028, 2029, 2030)}} \\
Full $-$ w/o residual refinement & $+8.035$ & $+6.209$ & $+7.635$ & $+7.293$ \\
Full $-$ learned gate & $-0.143$ & $-0.494$ & $+1.307$ & $+0.223$ \\
Full $-$ feature concatenation & $+2.366$ & $+1.350$ & $+2.283$ & $+2.000$ \\
Full $-$ CNN-only & $+1.173$ & $+0.610$ & $+1.735$ & $+1.173$ \\
Full $-$ sequence-only & $+10.909$ & $+9.638$ & $+11.312$ & $+10.620$ \\
Full $-$ both experts off & $+15.135$ & $+13.992$ & $+14.121$ & $+14.416$ \\
\bottomrule
\end{tabular}%
}
\end{table}

The paired analysis is used as a descriptive stability check rather than as a claim of definitive significance. Across the three matched seeds, the short full configuration consistently outperforms the zero-gate, w/o-distance, and late-fusion variants. The long full configuration likewise consistently outperforms the w/o-residual, feature-concatenation, sequence-only, and both-experts-off variants. The two native-length cross-configuration comparisons also retain the same direction for every seed. The correlation-removal and learned-gating controls instead quantify smaller changes relative to the main component and interface effects. The CNN-only control serves a different purpose: its larger gap from the complete long configuration supports complementary evidence from the state-space classifier, especially at low SNR.

\subsection{Accuracy--complexity trade-off}

Only the active configuration is instantiated for an experiment. Table~\ref{tab:complexity} reports the complete learned-operator estimate defined in Eq.~\eqref{eq:selective_scan_complexity}, rather than a tracer result that omits the fused scan. Latency is the median of 300 float32 batch-1 forward passes after 100 stabilization passes, measured with CUDA events in PyTorch 2.4.1. Peak allocated memory includes model parameters, the input, and inference activations. Both runtime quantities are measured on the RTX 4070 SUPER specified in Section~\ref{sec:experiments}.

\begin{table}[H]
\centering
\caption{Active-configuration complexity and batch-1 inference cost. FLOPs use the approximation $1\ \mathrm{MAC}\approx2$ scalar FLOPs.}
\label{tab:complexity}
\scriptsize
\resizebox{\textwidth}{!}{%
\begin{tabular}{lccccccc}
\toprule
Configuration & Length & Parameters (M) & MACs (M) & Approx. FLOPs (M) & Latency (ms) & Peak memory (MiB) & OAA (\%) \\
\midrule
Short-observation configuration & 128 & 0.330 & 11.13 & 22.25 & 2.85 & 10.51 & $67.279\pm0.135$ \\
Long-observation configuration & 1024 & 0.530 & 103.08 & 206.16 & 2.37 & 12.44 & $87.186\pm0.773$ \\
\bottomrule
\end{tabular}%
}
\end{table}

The long configuration requires approximately $9.3\times$ more MACs because of its longer feature sequence and residual branches. Its measured batch-1 latency is nevertheless slightly lower in this GPU setting. The short relation tokenizer contains several fine-grained descriptor and convolution kernels, whereas the longer operators expose more device parallelism; consequently, hardware latency is not proportional to the analytical MAC count. The latency and memory values are implementation- and hardware-specific, while parameter count and MACs provide the more portable comparison.

\section{Discussion}
\label{sec:discussion}

\subsection{Representation bottlenecks and expert design}

The ablations distinguish two design axes: what information an expert contributes and where that information enters the classifier. On the short RML2016.10b benchmark, explicit lagged distance contributes most of the relation branch's gain, and placing relation features before contextual encoding is more effective than combining a separate relation decision at the output. Correlation is retained as a complementary descriptor, while the distance channels provide the dominant measured gain. The near-constant nonnegative-SNR accuracy shows that the main relation-branch benefit is not a uniform increase in classifier capacity; it is concentrated where the available evidence is noisy.

On the long benchmark, multi-scale residual refinement acts on the feature map before tokenization, whereas the CNN expert collaborates with the sequence classifier at the decision level. Removing residual refinement produces a clear loss, concentrated below 0 dB, which supports feature correction as a distinct component of the long configuration. Fixed averaging performs comparably to learned gating and clearly exceeds feature concatenation. Because the learned gate adds a sample-dependent fusion module without a repeatable gain, the complete long configuration uses fixed averaging. When separately optimized as single-branch controls, the CNN-only model is stronger than the sequence-only model, and the complete model exceeds both. Within the jointly trained formal checkpoints, the larger residual coefficient at negative SNRs and the negatively correlated classifier margins further support adaptive feature correction and score-level collaboration. The long design thus separates feature correction from decision collaboration without requiring a learned fusion gate.

Cross-configuration results complement the component ablations. Across all seeds, the short configuration wins on RML2016.10b, while the long configuration wins on HisarMod2019. This consistency supports relation-token augmentation under limited observations and residual--classifier collaboration with extended support. The larger HisarMod2019 margin suggests that the latter uses richer temporal evidence more effectively. Within HisarMod2019, controlled windows further show that the Long--Short gap grows monotonically with temporal support for every seed. The long configuration thus scales more effectively with observation span on this data source. Its advantage at all four lengths, including 128 samples, also shows why nominal length alone is insufficient for assigning an expert configuration.

\subsection{Scope and limitations}

The two datasets differ in more than observation length. Their modulation sets, channel models, data distributions, and evaluation protocols differ simultaneously. The controlled HisarMod2019 study keeps sample identity and labels fixed while varying the retained temporal support, reducing this confounding and demonstrating a stable within-source trend. However, its nested windows are derived from fixed 1024-sample recordings rather than from independent acquisitions made at several durations. The evidence therefore shows more effective use of extended support by the long configuration on this data source; it does not isolate a causal rule based on sequence length alone.

The literature comparison in Table~\ref{tab:reported_comparison} uses reported results rather than same-codebase reimplementations; protocol and model-selection differences therefore limit strict cross-paper ranking. At the descriptor level, distance accounts for the dominant measured gain, whereas correlation is retained as a complementary descriptor.

\section{Conclusion}
\label{sec:conclusion}

This work introduced a unified backbone--expert framework for representation bottlenecks encountered under different AMR benchmark conditions. A common convolutional state-space backbone topology is paired with two complementary expert designs and integration interfaces. The short configuration adds lag-aware distance and complex-correlation tokens before sequence encoding, whereas the long configuration combines long-receptive-field residual refinement with sequence--CNN decision collaboration. The configurations achieved $67.28\pm0.14\%$ OAA on RML2016.10b and $87.19\pm0.77\%$ on HisarMod2019.

The short ablations show that the relation branch and its token interface mainly improve negative-SNR recognition, with distance providing the dominant measured contribution. On the long benchmark, residual refinement produces a clear gain, decision-level collaboration exceeds feature concatenation, and the CNN and sequence classifiers provide complementary evidence. The native-length cross-configuration comparison favors the configured design on both datasets, while the controlled HisarMod2019 study shows that the long configuration's advantage grows consistently with temporal support. Taken together, the results support a topology-unified design in which relation-token augmentation and residual--classifier collaboration address different representation bottlenecks.

\section*{CRediT authorship contribution statement}
Zhixiang Deng: Methodology, Software, Investigation, Validation, Formal analysis, Visualization, Writing--original draft. Houbiao Li: Conceptualization, Supervision, Writing--review and editing. Zongyong Cui: Validation, Writing--review and editing.

\section*{Declaration of competing interest}
The authors declare that they have no known competing financial interests or personal relationships that could have appeared to influence the work reported in this paper.

%\section*{Funding}
%This research received no external funding.

\section*{Data availability}
The RML2016.10b files used in this study are available from a public mirror on the \href{https://www.kaggle.com/datasets/marwanabudeeb/rml201610b}{Kaggle dataset page}. HisarMod2019 is available from the \href{https://ieee-dataport.org/open-access/hisarmod-new-challenging-modulated-signals-dataset}{IEEE DataPort dataset page}~\cite{oshea2016dataset,tekbiyik2020hisar}. The datasets are not redistributed by the authors. Code and experiment configurations will be made publicly available upon publication.

%\section*{Declaration of generative AI and AI-assisted technologies in the writing process}
%During the preparation of this work, the authors used OpenAI Codex and Google Gemini to assist with language %editing, code drafting and organization, and preliminary diagram ideation. All model implementations, experimental analyses, and final figures were reviewed and verified by the authors, and the final figures were manually redrawn. After using these tools, the authors reviewed and edited the content as needed and take full responsibility for the content of the publication.

\end{document}